\pdfoutput=1
\documentclass[11pt]{article}
\usepackage{acl2013}
\usepackage{times}
\usepackage{url}
\usepackage{latexsym}

\usepackage{stmaryrd}
\usepackage{amsmath}
\usepackage{mathtools,dcolumn}

\usepackage{tikz}
\usepackage{tikz-qtree}
\usetikzlibrary{arrows,positioning,calc,backgrounds,shapes,fit,snakes,mindmap,shadows}
\usetikzlibrary{matrix,decorations.pathreplacing}

\usepackage{subcaption}
\usepackage{multirow}

\newcommand\km[1]{}
\newcommand\ed[1]{}

\tikzstyle{vector}=[rectangle,inner sep=2pt,label distance=4pt,align=right,rounded corners,draw,font=\footnotesize]

\newcommand{\pvector}[1]{$\begin{pmatrix}#1\end{pmatrix}$}

\title{``Not not bad'' is not ``bad'': A distributional account of negation}

\author{Karl Moritz Hermann \\
  \And
  Edward Grefenstette \\
  University of Oxford Department of Computer Science \\
  Wolfson Building, Parks Road \\
  Oxford OX1 3QD, United Kingdom \\
  {\tt firstname.lastname@cs.ox.ac.uk} \\\And
  Phil Blunsom \\
  }

\date{}

\begin{document}
\maketitle
\begin{abstract}
With the increasing empirical success of distributional models of compositional semantics, it is timely to consider the types of textual logic that such models are capable of capturing.
In this paper, we address shortcomings in the ability of current models to capture logical operations such as negation.
As a solution we propose a tripartite formulation for a continuous vector space representation of semantics and subsequently use this representation to develop a formal compositional notion of negation within such models.
\end{abstract}

\section{Introduction} 
\label{sec:introduction}

Distributional models of semantics characterize the meanings of words as a function of the words they co-occur with \cite{Firth1957}. These models, mathematically instantiated as sets of vectors in high dimensional vector spaces, have been applied to tasks such as thesaurus extraction \cite{Grefenstette1994,Curran2004}, word-sense discrimination \cite{schutze1998automatic}, automated essay marking \cite{Landauer1997}, and so on. 


During the past few years, research has shifted from using distributional methods for modelling the semantics of words to using them for modelling the semantics of larger linguistic units such as phrases or entire sentences. This move from word to sentence has yielded models applied to tasks such as paraphrase detection \cite{mitchell2008vector,Mitchell2010,Grefenstette:2011,blacoe70comparison}, sentiment analysis \cite{socherEMNLP12,Hermann:2013}, and semantic relation classification (\emph{ibid.}). Most efforts approach the problem of modelling phrase meaning through vector composition using linear algebraic vector operations \cite{mitchell2008vector,Mitchell2010,zanzotto2010estimating}, matrix or tensor-based approaches \cite{Baroni2010,Coecke2010,grefenstette2013multistep,Kartsaklis2012}, or through the use of recursive auto-encoding \cite{socher2011dynamic,Hermann:2013} or neural-networks \cite{socherEMNLP12}. On the non-compositional front, \newcite{Erk2008} keep word vectors separate, using syntactic information from sentences to disambiguate words in context; likewise \newcite{turney2012domain} treats the compositional aspect of phrases and sentences as a matter of similarity measure composition rather than vector composition.

These compositional distributional approaches often portray themselves as attempts to reconcile the empirical aspects of distributional semantics with the structured aspects of formal semantics. However, they in fact only principally co-opt the syntax-sensitivity of formal semantics, while mostly eschewing the logical aspects.

Expressing the effect of logical operations in high dimensional distributional semantic models is a very different task than in boolean logic. For example, whereas predicates such as `red' are seen in predicate calculi as functions mapping elements of some set $M_{red}$ to $\top$ (and all other domain elements to $\bot$), in compositional distributional models we give the meaning of `red' a vector-like representation, and devise some combination operation with noun representations to obtain the representation for an adjective-noun pair. Under the logical view, negation of a predicate therefore yields a new truth-function mapping elements of the complement of $M_{red}$ to $\top$ (and all other domain elements to $\bot$), but the effect of negation and other logical operations in distributional models is not so sharp: we expect the representation for ``not red'' to remain close to other objects of the same domain of discourse (i.e.~other colours) while being sufficiently different from the representation of `red' in some manner. 
Exactly how textual logic would best be represented in a continuous vector space model remains an open problem.

In this paper we propose one possible formulation for a continuous vector space based representation of semantics.
We use this formulation as the basis for providing an account of logical operations for distributional models.
In particular, we focus on the case of negation and how it might work in higher dimensional distributional models.
Our formulation separates domain, value and functional representation in such a way as to allow negation to be handled naturally.
We explain the linguistic and model-related impacts of this mode of representation and discuss how this approach could be generalised to other semantic functions. 

In Section~\ref{sec:related_work}, we provide an overview of work relating to that presented in this paper, covering the integration of logical elements in distributional models, and the integration of distributional elements in logical models.
In Section~\ref{sec:logic_in_text}, we introduce and argue for a tripartite representation in distributional semantics, and discuss the issues relating to providing a linguistically sensible notion of negation for such representations.
In Section~\ref{sec:a_general_matrix_vector_model}, we present matrix-vector models similar to that of~\newcite{socherEMNLP12} as a good candidate for expressing this tripartite representation. We argue for the elimination of non-linearities from such models, and thus show that negation cannot adequately be captured.
In Section~\ref{sec:analysis}, we present a short analysis of the limitation of these matrix-vector models with regard to the task of modelling non-boolean logical operations, and present an improved model bypassing these limitations in Section~\ref{sec:an_improved_model}. Finally, in Section~\ref{sec:conclusions_and_further_work}, we conclude by suggesting future work which will extend and build upon the theoretical foundations presented in this paper.

         
\section{Motivation and Related Work} 
\label{sec:related_work}

The various approaches to combining logic with distributional semantics can broadly be put into three categories: those approaches which use distributional models to enhance existing logical tools; those which seek to replicate logic with the mathematical constructs of distributional models; and those which provide new mathematical definitions of logical operations within distributional semantics. The work presented in this paper is in the third category, but in this section we will also provide a brief overview of related work in the other two in order to better situate the work this paper will describe in the literature.

\paragraph{Vector-assisted logic} The first class of approaches seeks to use distributional models of word semantics to enhance logic-based models of textual inference. The work which best exemplifies this strand of research is found in the efforts of \newcite{garrette2011integrating} and, more recently, \newcite{beltagy:starsem13}. This line of research converts logical representations obtained from syntactic parses using Bos' Boxer \cite{bos2008wide} into Markov Logic Networks \cite{richardson2006markov}, and uses distributional semantics-based models such as that of \newcite{Erk2008} to deal with issues polysemy and ambiguity. 

As this class of approaches deals with improving logic-based models rather than giving a distributional account of logical function words, we view such models as orthogonal to the effort presented in this paper.

\paragraph{Logic with vectors} The second class of approaches seeks to integrate boolean-like logical operations into distributional semantic models using existing mechanisms for representing and composing semantic vectors. \newcite{Coecke2010} postulate a mathematical framework generalising the syntax-semantic passage of Montague Grammar~\cite{Montague1974} to other forms of syntactic and semantic representation. They show that the parses yielded by syntactic calculi satisfying certain structural constraints can be canonically mapped to vector combination operations in distributional semantic models. They illustrate their framework by demonstrating how the truth-value of sentences can be obtained from the combination of vector representations of words and multi-linear maps standing for logical predicates and relations. They furthermore give a matrix interpretation of negation as a `swap' matrix which inverts the truth-value of vectorial sentence representations, and show how it can be embedded in sentence structure. 

Recently, \newcite{Grefenstette2013TFDS} showed that the examples from this framework could be extended to model a full quantifier-free predicate logic using tensors of rank 3 or lower. In parallel, \newcite{socherEMNLP12} showed that propositional logic can be learned using tensors of rank 2 or lower (i.e.~only matrices and vectors) through the use of non-linear activation functions in recursive neural networks.

The work of~\newcite{Coecke2010} and~\newcite{Grefenstette2013TFDS} limits itself to defining, rather than learning, distributional representations of logical operators for distributional models that simulate logic, and makes no pretense to the provision of operations which generalise to higher-dimensional distributional semantic representations.
As for the non-linear approach of \newcite{socherEMNLP12}, we will discuss, in Section~\ref{sec:a_general_matrix_vector_model} below, the limitations with this model with regard to the task of modelling logic for higher dimensional representations.

\paragraph{Logic for vectors} The third and final class of approaches is the one the work presented here belongs to. This class includes attempts to define representations for logical operators in high dimensional semantic vector spaces. Such approaches do not seek to retrieve boolean logic and truth values, but to define what logical operators \emph{mean} when applied to distributional representations. The seminal work in this area is found in the work of \newcite{widdows2003word}, who define negation and other logical operators algebraically for high dimensional semantic vectors. Negation, under this approach, is effectively a binary relation rather than a unary relation: it expresses the semantics of statements such as `A NOT B' rather than merely `NOT B', and does so by projecting the vector for A into the orthogonal subspace of the vector for B. This approach to negation is useful for vector-based information retrieval models, but does not necessarily capture all the aspects of negation we wish to take into consideration, as will be discussed in Section~\ref{sec:logic_in_text}.


\section{Logic in text}
\label{sec:logic_in_text}



In order to model logical operations over semantic vectors, we propose a tripartite meaning representation, which combines the separate and distinct treatment of domain-related and value-related aspects of semantic vectors with a domain-driven syntactic functional representation. This is a unification of various recent approaches to the problem of semantic representation in continuous distributional semantic modelling \cite{socherEMNLP12,turney2012domain,Hermann:2013}.

We borrow from~\newcite{socherEMNLP12} and others \cite{Baroni2010,Coecke2010} the idea that the information words refer to is of two sorts: first the semantic content of the word, which can be seen as the sense or reference to the concept the word stands for, and is typically modelled as a semantic vector; and second, the \emph{function} the word has, which models the effect the word has on other words it combines with in phrases and sentences, and is typically modelled as a matrix or higher-order tensor. We borrow from~\newcite{turney2012domain} the idea that the semantic aspect of a word should not be modelled as a single vector where everything is equally important, but ideally as two or more vectors (or, as we do here, two or more \emph{regions} of a vector) which stand for the aspects of a word relating to its \emph{domain}, and those relating to its \emph{value}. 

We therefore effectively suggest a \emph{tripartite representation} of the semantics of words: a word's meaning is modelled by elements representing its value, domain, and function, respectively.

\paragraph{The tripartite representation} 
We argue that the tripartite representation suggested above allows us to explicitly capture several aspects of semantics. 
Further, while there may be additional distinct aspects of semantics, we argue that this is a minimal viable representation.

First of all, the differentiation between domain and value is useful for establishing similarity within subspaces of meaning.
For instance, the words \textit{blue} and \textit{red} share a common domain (colours) while having very different values. We hypothesise that making this distinction explicit will allow for the definition of more sophisticated and fine-grained semantics of logical operations, as discussed below. Although we will represent domain and value as two regions of a vector, there is no reason for these not to be treated as separate vectors at the time of comparison, as done by~\newcite{turney2012domain}.

Through the third part, the functional representation, we capture the compositional aspect of semantics: the functional representation governs how a term interacts with its environment. Inspired by the distributional interpretation \cite{Baroni2010,Coecke2010} of syntactically-paramatrized semantic composition functions from Montogovian semantics~\cite{Montague1974}, we will also assume the function part of our representation to be parametrized principally by syntax and domain rather than value. The intuition behind taking domain into account in addition to syntactic class being that all members of a domain largely interact with their environment in the same fashion. 



\paragraph{Modeling negation} The tripartite representation proposed above allows us to define logical operations in more detail than competing approaches. To exemplify this, we focus on the case of negation. 

We define negation for semantic vectors to be the absolute complement of a term in its domain.
This implies that negation will not affect the domain of a term but only its value.
Thus, \textit{blue} and \textit{not blue} are assumed to share a common domain. We call this naive form of negation the inversion of a term A, which we idealise as the partial inversion $A^{inv}$ of the region associated with the value of the word in its vector representation $A$.

\begin{figure}[ht]
\begin{align*} 
\begin{matrix}
  \left[\begin{array}{c}d\\v\\v\end{array}\right] &
  \left[\begin{array}{c}d\\v\\-v\end{array}\right] &  
  \left[\begin{array}{c}d\\v\\-\mu v\end{array}\right] \\[1.5em]
  \Bigl[\begin{array}{>{\centering\arraybackslash$} p{0.7em} <{$}}f\end{array}\Bigr] &  
  \Bigl[\begin{array}{>{\centering\arraybackslash$} p{1.5em} <{$}}f\end{array}\Bigr] &
  \Bigl[\begin{array}{>{\centering\arraybackslash$} p{2.1em} <{$}}f\end{array}\Bigr] \\[1.3em]
\text{W} &
\text{W}^{inv} &
\neg\text{W}
\end{matrix}
\end{align*}

\caption{The semantic representations of a word $W$, its inverse $W^{inv}$ and its negation $\neg W$. The domain part of the representation remains unchanged, while the value part will partially be inverted (inverse), or inverted and scaled (negation) with $0 < \mu < 1$. The (separate) functional representation also remains unchanged.}\label{fig:negation}
\end{figure}

Additionally, we expect negation to have a diminutive effect. This diminutive effect is best exemplified in the case of sentiment: \textit{good} is more positive than \textit{not bad}, even though \textit{good} and \textit{bad} are antonyms of each other. By extension \textit{not not good} and \textit{not not not bad} end up somewhere in the middle---qualitative statements still, but void of any significant polarity. To reflect this diminutive effect of negation and double negation commonly found in language, we define the idealised diminutive negation $\lnot A$ of a semantic vector $A$ as a scalar inversion over a segment of the value region of its representation with the scalar $\mu: 0 < \mu < 1$, as shown in Figure~\ref{fig:negation}.




As we defined the functional part of our representation to be predominately parametrized by syntax and domain, it will remain constant under negation and inversion.

\section{A general matrix-vector model}
\label{sec:a_general_matrix_vector_model}

Having discussed, above, how the vector component of a word can be partitioned into domain and value, we now turn to the partition between semantic content and function. A good candidate for modelling this partition would be a dual-space representation similar to that of~\newcite{socherEMNLP12}. In this section, we show that this sort of representation is not well adapted to the modelling of negation.

Models using dual-space representations have been proposed in several recent publications, notably in \newcite{turney2012domain} and \newcite{socherEMNLP12}.
We use the class of recursive matrix-vector models as the basis for our investigation; for a detailed introduction see the MV-RNN model described in \newcite{socherEMNLP12}.

We begin by describing composition for a general dual-space model, and apply this model to the notion of compositional logic in a tripartite representation discussed earlier.
We identify the shortcomings of the general model and subsequently discuss alternative composition models and modifications that allow us to better capture logic in vector space models of meaning.


Assume a basic model of compositionality for such a tripartite representation as follows. 
Each term is encoded by a semantic vector $v$ capturing its domain and value, as well as a matrix $M$ capturing its function.
Thus, composition consists of two separate operations to learn semantics and function of the composed term:
\begin{align}
\mathbf{v}_p &= f_{v}(\mathbf{v}_a,\mathbf{v}_b,M_a,M_b)\\
M_p &= f_{M}(M_a,M_b)\nonumber
\end{align}
As we defined the functional representation to be parametrized by syntax and domain, its composition function does not require $\mathbf{v}_a$ and $\mathbf{v}_b$ as inputs, with all relevant information already being contained in $M_a, M_b$.
In the case of \newcite{socherEMNLP12} these functions are as follows:
\begin{align}
M_p &= W_M\begin{bmatrix}M_a\\M_b\end{bmatrix}\\
\mathbf{v}_p &= g\left(W_v\begin{bmatrix}M_a \mathbf{v}_b\\M_b \mathbf{v}_a\end{bmatrix}\right)
\end{align}
where $g$ is a non-linearity.

\subsection{The question of non-linearities}

While the non-linearity $g$ could be equipped with greater expressive power, such as in the boolean logic experiment in \newcite{socherEMNLP12}), the aim of this paper is to place the burden of compositionality on the atomic representations instead. For this reason we treat $g$ as an identity function, and $W_M$, $W_v$ as simple additive matrices in this investigation, by setting
\[
g = I \quad W_v = W_M = \left[I\ I\right]
\]
where $I$ is an identity matrix. This simplification is justified for several reasons.

A simple non-linearity such as the commonly used hyperbolic tangent or sigmoid function will not add sufficient power to overcome the issues outlined in this paper.
Only a highly complex non-linear function would be able to satisfy the requirements for vector space based logic as discussed above.
Such a function would defeat the point however, by pushing the ``heavy-lifting'' from the model structure into a separate function.

Furthermore, a non-linearity effectively encodes a scattergun approach:
While it may have the power to learn a desired behaviour, it similarly has a lot of power to learn undesired behaviours and side effects.
From a formal perspective it would therefore seem more prudent to explicitly encode desired behaviour in a model's structure rather than relying on a non-linearity.


\subsection{Negation}
\begin{figure}[b]
{ \centering

\begin{center}
  \pvector{
1   &       &       &       &     &     \\
    & \ddots    &       &     &\mathbf{0} &   \\
    &       & 1     &\\
    &       &     & -\mu \\
    &\mathbf{0} &     &       & \ddots      \\
    &     &     &       &     & -\mu   
}
\end{center}

}
\caption{A partially scaled and inverted identity matrix $J_\mu$. Such a matrix can be used to transform a vector storing a domain and value representation into one containing the same domain but a partially inverted value, such as $W$ and $\neg W$ described in Figure~\ref{fig:negation}.}\label{fig:partialident}
\end{figure}

We have outlined our formal requirements for negation in the previous section.
From these requirements we can deduce four equalities, concerning the effect of negation and double negation on the semantic representation and function of a term. The matrices $J_\mu$ and $J_\nu$ (illustrated in Figure~\ref{fig:partialident}) describe a partially scaled and inverted identity matrix, where $0<\mu, \nu < 1$.
\begin{align}
f_v(\text{not},a) &= J_\mu \mathbf{v}_a \label{eq:neg-0} \\
f_M(\text{not},a) &\approx M_a \label{eq:neg-1} \\
f_v(\text{not},f_v(\text{not},a)) &= J_\nu J_\mu \mathbf{v}_a \\
f_M(\text{not},f_M(\text{not},a)) &\approx M_a \label{eq:neg-3}
\end{align}

Based on our assumption about the constant domain and interaction across negation, we can replace the approximate equality with a strict equality in Equations~\ref{eq:neg-1} and~\ref{eq:neg-3}.
Further, we assume that both $M_a \neq I$ and $M_a \neq 0$, i.e. that $A$ has a specific and non-zero functional representation. 
We make a similar assumption for the semantic representation $\mathbf{v}_a \neq 0$.

Thus, to satisfy the equalities in Equations~\ref{eq:neg-0} through~\ref{eq:neg-3}, we can deduce the values of $\mathbf{v}_{not}$ and $M_{not}$ as discussed below.

\paragraph{Value and Domain in Negation}

Under the simplifications of the model explained earlier, we know that the following is true:
\begin{align*}
f_v(a,b)  &= g\left(W_v\begin{bmatrix}M_a \mathbf{v}_b\\M_b \mathbf{v}_a\end{bmatrix}\right)\\
    &= I\left(\begin{bmatrix}I\ I\end{bmatrix}\begin{bmatrix}M_a \mathbf{v}_b\\M_b \mathbf{v}_a\end{bmatrix}\right)\\
    &= M_a \mathbf{v}_b + M_b \mathbf{v}_a
\end{align*}
I.e. the domain and value representation of a parent is the sum of the two $Mv$ multiplications of its children.
The matrix $W_v$ could re-weight this addition, but that would not affect the rest of this analysis.

Given the idea that the domain stays constant under negation and that a part of the value is inverted and scaled, we further know that these two equations hold:
\begin{align*}
&\forall a\in A: f_v(\text{not},a) = J_\mu \mathbf{v}_a \\
&\forall a\in A: f_v(\text{not},f_v(\text{not},a)) = J_\nu J_\mu \mathbf{v}_a
\end{align*}

Assuming that both semantic and functional representation across all $A$ varies and is non-zero, these equalities imply the following conditions for the representation of \textit{not}:
\begin{align*}
M_{not} &= J_\mu = J_\nu \\
\mathbf{v}_{not} &= 0
\end{align*}
These two equations suggest that the term not has no inherent value ($\mathbf{v}_{not} = 0$), but merely acts as a function, inverting part of another terms semantic representation ($M_{not} = J_\mu$).

\paragraph{Functional Representation in Negation}
We can apply the same method to the functional representation.
Here, we know that:
\begin{align*}
f_M(a,b)  &= W_M\begin{bmatrix}M_a\\M_b\end{bmatrix}\\
    &= \begin{bmatrix}I\ I\end{bmatrix}\begin{bmatrix}M_a\\M_b\end{bmatrix}\\
    &= M_a + M_b
\end{align*}

Further, as defined in our discussion of negation, we require the functional representation to remain unchanged under negation:
\begin{align*}
\forall a\in A: f_M(\text{not},a) &= M_a \\
\forall a\in A: f_M(\text{not},f_M(\text{not},a)) &= M_a
\end{align*}
These requirements combined leave us to conclude that $M_{not} = 0$.
Combined with the result from the first part of the analysis, this causes a contradiction:
\begin{align*}
&M_{not} = 0\\
&M_{not} = J_\mu\\
\implies &J_\mu = 0 \lightning
\end{align*}

This demonstrates that the MV-RNN as described in this paper is not capable of modelling semantic logic according to the principles we outlined.
The fact that we would require $M_{not} = 0$ further supports the points made earlier about the non-linearities and setting $W_M$ to $\begin{bmatrix}I\ I\end{bmatrix}$.
Even a specific $W_M$ and non-linearity would not be able to ensure that the functional representation stays constant under negation given a non-zero $M_{not}$.

Clearly, any other complex semantic representation would suffer from the same issue here---the failure of double-negation to revert a representation to its (diminutive) original.

\section{Analysis}
\label{sec:analysis}

The issue identified with the MV-RNN style models described above extends to a number of other models of vector spaced compositionality.
It can be viewed as a problem of uninformed composition caused by a composition function that fails to account for syntax and thus for scope.

\begin{figure}[t]
\centering
\begin{tikzpicture}
\Tree
[.S
	[.NP [.Det This ] [.N car ] ]
	[.VP
		[.VBZ is ]
		[.RB not ]
		[.ADJP [.JJ blue ] ]
	]
]
\end{tikzpicture}
\caption{The parse tree for \textit{This car is not blue}, highlighting the limited scope of the negation.}\label{fig:neg-scope}
\end{figure}
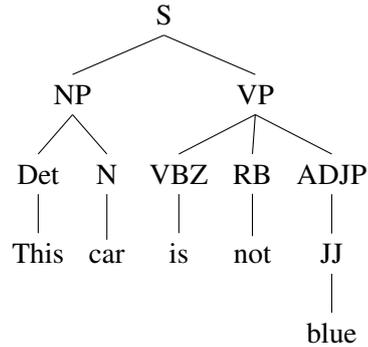

Of course, identifying the scope of negation is a hard problem in its own right---see e.g. the *SEM 2012 shared task \cite{Morante:2012}.
However, at least for simple cases, we can deduce scope by considering the parse tree of a sentence:
If we consider the parse tree for \textit{this car is not blue}, it is clear that the scope of the negation expressed includes the colour but not the car (Figure~\ref{fig:neg-scope}).

While the MV-RNN model in \newcite{socherEMNLP12} incorporates parse trees to guide the order of its composition steps, it uses a single composition function across all steps.
Thus, the functional representation of \textit{not} will to some extent propagate outside of its scope, leading to a vector capturing something that is not blue, but also not quite a car.

There are several possibilities for addressing this issue.
One possibility is to give greater weight to syntax, for instance by parametrizing the composition functions $f_v$ and $f_M$ on the parse structure.
This could be achieved by using specific weight matrices $W_v$ and $W_M$ for each possible tag.
While the power of this approach is limited by the complexity of the parse structure, it would be better able to capture effects such as the scoping and propagation of functional representations.

Another approach, which we describe in greater detail in the next section, pushes the issue of propagation onto the word level.
While both approaches could easily be combined, this second option is more consistent with our aim of avoiding the implicit encoding of logic into fixed model parameters in favour of the explicit encoding in model structure.


\section{An improved model}
\label{sec:an_improved_model}

As we outlined in this paper, a key requirement for a compositional model motivated by formal semantics is the ability to propagate functional representations, but also to not propagate these representations when doing so is not semantically appropriate.
Here, we propose a modification of the MV-RNN class of models that can capture this distinction without the need to move the composition logic into the non-linearity.

We add a parameter $\alpha$ to the representation of each word, controlling the degree to which its functional representation propagates after having been applied in its own composition step.

Thus, the composition step of the new model requires three equations:
\begin{align}
M_p 			&= W_M\begin{bmatrix}\frac{\alpha_a}{\alpha_a+\alpha_b} M_a\\\frac{\alpha_b}{\alpha_a+\alpha_b} M_b\end{bmatrix}\\[0.4em]
\mathbf{v}_p 			&= g\left(W_v\begin{bmatrix}M_a \mathbf{v}_b\\M_b \mathbf{v}_a\end{bmatrix}\right)\\[0.4em]
\alpha_p 	&= max(\alpha_a,\alpha_b)
\end{align}

Going back to the discussion of negation, this model has the clear advantage of being able to capture negation in the way we defined it.
As $f_v(a,b)$ is unchanged, these two equations still hold:
\begin{align*}
M_{not} &= J_\mu = J_\nu \\
\mathbf{v}_{not} &= 0
\end{align*}
However, as $f_M(a,b)$ is changed, the second set of equations changes.
We use $Z$ as the $\alpha$-denominator ($Z = \alpha_a + \alpha_B$) for simplification:
\begin{align*}
f_M(a,b)  &= W_M\begin{bmatrix}\frac{\alpha_a}{Z} M_a\\\frac{\alpha_b}{Z} M_b\end{bmatrix}\\
    &= \begin{bmatrix}I\\I\end{bmatrix}\begin{bmatrix}\frac{\alpha_a}{Z} M_a\\\frac{\alpha_b}{Z} M_b\end{bmatrix}\\
    &= \frac{\alpha_a}{Z} M_a + \frac{\alpha_b}{Z} M_b
\end{align*}
Further, we still require the functional representation to remain constant under negation:
\begin{align*}
\forall a\in A: f_M(\text{not},a) &= M_a \\
\forall a\in A: f_M(\text{not},f_M(\text{not},a)) &= M_a
\end{align*}
Thus, we can infer the following two conditions on the new model:
\begin{align*}
\frac{\alpha_{not}}{Z} M_{not} \approx 0\\
\frac{\alpha_a}{Z} M_a \approx M_{a}
\end{align*}
From our previous investigation we already know that $M_{not} = J_\mu \neq 0$, i.e. that \textit{not} has a non-zero functional representation.
While this caused a contradiction for the original MV-RNN model, the design of the improved model can resolve this issue through the $\alpha$-parameter:
\begin{align*}
\alpha_{not} = 0
\end{align*}
Thus, we can use this modified MV-RNN model to represent negation according to the principles outlined in this paper.
The result $\alpha_{not} = 0$ is in accordance with our intuition about the propagation of functional aspects of a term:
We commonly expect negation to directly affect the things under its scope (\textit{not blue}) by choosing their semantic complement.
However, this behaviour should not propagate outside of the scope of the negation.
A \textit{not blue car} is still very much a car, and when a film is not good, it is still very much a film.

\section{Discussion and Further Work} 
\label{sec:conclusions_and_further_work}


In this paper, we investigated the capability of continuous vector space models to capture the semantics of logical operations in non-boolean cases. Recursive and recurrent vector models of meaning have enjoyed a considerable amount of success in recent years, and have been shown to work well on a number of tasks.
However, the complexity and subsequent power of these models comes at the price that it can be difficult to establish which aspect of a model is responsible for what behaviour.
This issue was recently highlighted by an investigation into recursive autoencoders for sentiment analysis \cite{Scheible:2013}. Thus, one of the key challenges in this area of research is the question of how to control the power of these models.
This challenge motivated the work in this paper.
By removing non-linearities and other parameters that could disguise model weaknesses, we focused our work on the basic model design.
While such features enhance model power, they should not be used to compensate for inherently flawed model designs.

As a prerequisite for our investigation we established a suitable encoding of textual logic.
Distributional representations have been well explained on the word level, but less clarity exists as to the semantic content of compositional vectors.
With the tripartite meaning representation we proposed one possible approach in that direction, which we subsequently expanded by discussing how negation should be captured in this representation.

Having established a suitable and rigorous system for encoding meaning in compositional vectors, we were thus able to investigate the representative power of the MV-RNN model. 
We focused this paper on the case of negation, which has the advantage that it does not require many additional assumptions about the underlying semantics.
Our investigation showed that the basic MV-RNN model is incompatible with our notion of negation and thus with any textual logic building on this proposal.

Subsequently, we analysed the reasons for this failure.
We explained how the issue of negation affects the general class of MV-RNN models.
Through the issue of double-negation we further showed how this issue is largely independent on the particular semantic encoding used.
Based on this analysis we proposed an improved model that is able to capture such textual logic.

In summary, this paper has two key contributions.
First, we developed a tripartite representation for vector space based models of semantics, incorporating multiple previous approaches to this topic.
Based on this representation, the second contribution of this paper was a modified MV-RNN model that can capture effects such as negation in its inherent structure.

In future work, we would like to build on the proposals in this paper, both by extending our work on textual logic to include formulations for e.g. function words, quantifiers, or locative words.
Similarly, we plan to experimentally validate these ideas.
Possible tasks for this include sentiment analysis and relation extraction tasks such as in \newcite{socherEMNLP12} but also more specific tasks such as the *SEM shared task on negation scope and reversal \cite{Morante:2012}.

\section*{Acknowledgements}

The first author is supported by the UK Engineering and Physical Sciences Research Council (EPSRC). The second author is supported by EPSRC Grant \texttt{EP/I03808X/1}.

\bibliographystyle{acl}
\bibliography{a,logicbiblio}
\end{document}